# Radial-Layer Jamming Mechanism for String Configuration

Rio Mukaide*, Kenjiro Tadakuma**, Masahiro Watanabe*, Yu ozawa*, Tomoya Takahashi*, Masashi Konyo*, Satoshi Tadokoro*

*Abstract*—Grippers can be attached to objects in a rigid mode, and they are therefore used in various applications, for example granular jamming gripper. This paper introduces a cutting-edge radial layer jamming mechanism with is tunable stiffness, which is critical for the development of grippers. The layer jamming mechanism generates friction between the layers of multi cylindrical walls by pulling wire. This paper describes the principles of three types of proposed tendon-driven jamming mechanism, in addition to their prototypes of string configuration and the experiments conducted on the holding torques of their joints. Due to the string configuration, the surface and three-dimensional (3D) shape. This mechanism can be implemented in various applications.

## I. Introduction

The proposed mechanism with tunable stiffness is useful for grippers. The grippers attach to the form of an object and grip it in a rigid mode; such as the soft gripper [1], which employs a tendon drive. Moreover, several mechanisms employ tendon drives, such as the bio-inspired manipulator [2], and tendon-driven continuum manipulators [3]. At present, there are various types of mechanisms with tunable stiffnesses, such as the granular jamming [4] [5] and layer jamming mechanisms. The universal robotic gripper [6], which is a variable stiffness robotic gripper based on integrated soft actuation and particle jamming [7], represents a jamming mechanism that employs granular material. By evacuating the space has particles, those particles contact with each other and the friction is originated between them. It is called a granular jamming. This is referred to as holding force. Moreover, there is a layer jamming that involves the generation of friction by the forced contact thin-plate laminates. For example "utilizing layer jamming" [8] [9] that uses silicone rubber as a case store layer and press the layer by evacuating it, and "Selective Stiffening of Soft Actuators Based on Jamming" [10]. However, the abovementioned methods are disadvantageous. First, they cannot output higher torques when the air pressure in the case reaches 0 Pa. Second, if the silicone rubber is rent, friction is not generated between layers.

*Rio Mukaide, Kenjiro Tadakuma, Masahiro Watanabe, Yu Ozawa, Tomoya Takahashi, Masashi Konyo and Satoshi Tadokoro are with the Graduate school of Information Sciences, Tohoku University, Japan (email: mukaide.rio@rm.is.tohoku.ac.jp).

In this paper, a string configuration jamming mechanism, which can increase its holding torques with respect to the tensile load using a tendon drive, is proposed. First, a novel flexible string configuration that can realize significant holding torque was developed. Second, the shape of the bead joints was clarified as a string configuration unit. This mechanism has three characteristics. First, the robust architecture was realized using rigid-body parts. Second, the mechanism can be used for the development of facies and three-dimensional (3D) shapes, given that the mechism can be considered as a "line" element. Third, there is no significant limitation with respect to material use, that structures can be composited using heat and chemical resistant materials.

In this study, compositions were developed using wire and beads as a string configuration, to achieve a tunable stiffness; in addition to moveable joints of the multilayer structure. In addition, beads that incorporated a composition of the string configuration and joints were manufactured. The experiment on the measurement of the holding torque of joints was then conducted to confirm its effectiveness.

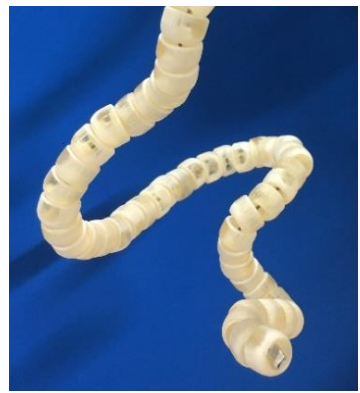

Fig. 1. Bead jamming mechanism prototype

## II. Principle

Three types of mechanisms were exploited for the development of the string configuration with tunable stiffness, using wire. As can be seen in Fig. 2, the beads jamming mechanism does not contain layers, and only involves one contact surface. The two other types are considered as layer jamming mechanisms, which increase the friction between the layer, and can increase the holding torque. Two types of laminating directions are possible: the z-axis or r-axis. The comb jamming mechanism is a z-axis type, and the radial-layer is an r-axis type that involves laminated concentric circles.

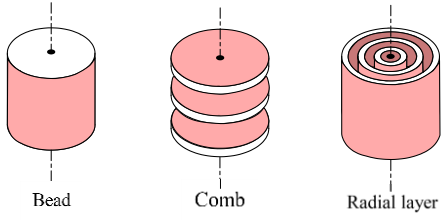

Fig. 2. Three types of jamming mechanisms (bead, comb, and radial-layer)

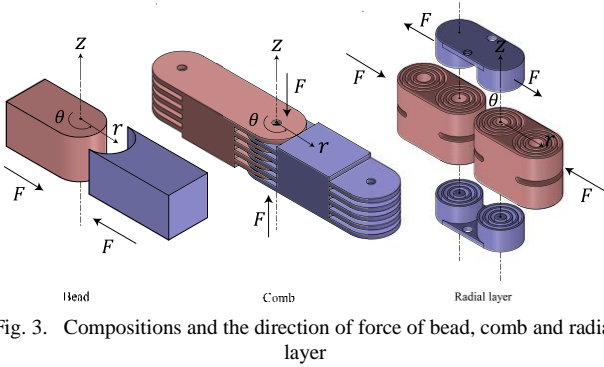

Fig. 3. Compositions and the direction of force of bead, comb and radial layer

*A. Bead Jamming*

The bead jamming mechanism involves only one contact surface, which is the cylindrical surface. In this case, there is a joint that fits into the hollow, as shown in Fig. 3. Moreover, as shown in Fig. 4(b), they are coupled by the wire. When the wire is pulled, they are forced into contact with each other, and the friction originated can maintain the shapes of the string configuration. The highly-articulated robotic probe for minimally invasive surgery developed by Degani and Choset is based on the same mechanism [11]. However, its holding torque is dependent on a low surface frictional coefficient; thus, so the torque increases slightly in accordance with an increase in the tensile load. Therefore, for a higher holding torque, the radii of the beads should be increased.

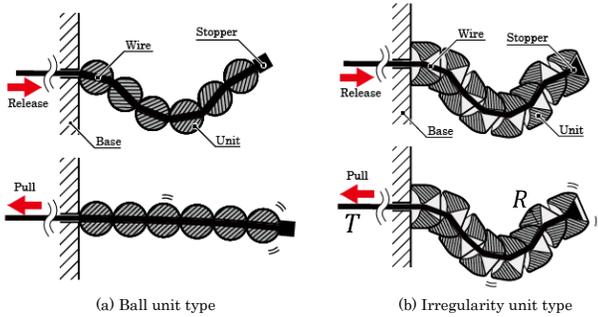

Fig. 4. Principle of bead jamming

The theoretical holding torque for the bead jamming mechanism is as follows:

$$\tau_{\text{beads}} = \mu R T \qquad (1)$$

when the frictional coefficient of the surface is µ, the radius of beads is $R$, and tensile load is $T$.

In Fig. 4(a), a simple ball with a through-hole cannot maintain its curved string configuration, and its only form is that of a straight line when the wire is pulled. Nevertheless, the composition with its fitting structure and conical wire route, as shown in Fig. 4(b), can help maintain its form, given that the length of the wire route remains almost unchanged from its previous state.

*B. Comb Jamming*

From a comparison with the comb jamming mechanism, which can output higher torques than the bead jamming mechanism; a comb jamming mechanism was developed, wherein thin plates are laminated to the z-axis, as shown in Fig. 2. Moreover, the number of frictional surfaces can be increased by incorporating comb parts into the bead jamming mechanism (Fig. 3). Friction is generally independent of the contact area [12]; however, the same normal load is added to each plate in the case of this mechanism. Hence, the total friction increases linearly in accordance with an increase in the number of plates, and a significantly holding torque can be realized.

The holding torque of the joints was derived under the assumption that the plates are not subject to bending. The total normal load is $F_n$. If the contact area is considered a round shape with radius $R$, the minute normal load $dF_n$ for the layer can be expressed as follows:.

$$dF_n = F_n \cdot \frac{r\, dr\, d\theta}{\pi R^2} \qquad (2)$$

The minute holding torque $d\tau$ that is applied to the small area can be expressed as follows:

$$d\tau = \mu \cdot dF_n \cdot r \qquad (3)$$

The holding torque that occurs in a frictional surface can be expressed as follows:

$$\int d\tau = \int \mu r\, dF_n \\
= \int_0^{2\pi} \int_0^R \frac{\mu F_n}{\pi R^2} r^2\, dr\, d\theta \\
= \frac{2}{3} \pi \mu F_n R \qquad (4)$$

Moreover, the torque is experienced by each contact surface. The total holding torque can then be expressed as follows, where $a$ is the number of plates.

$$\tau_{\text{comb}} = \frac{2(2a-1)}{3} \pi \mu F_n R \qquad (5)$$

This mechanism can output a significantly high torque. However, it requires the pulley to convert tensile loads to normal loads towards the normal surface of the layer. Moreover, two challenges were noted. First, the width of the string configuration was excessively long. Second, the rolling friction [13] [14] between the pulley and wire gradually decreased the tension of the wire from the root of the string configuration to its tip. Consequently, the holding torque of its tip decreased.

The comb jamming mechanism differs from the bead jamming mechanism with respect to its responsiveness. In the case of the bead jamming mechanism, surfaces can instantly come into contact by pulling the wire, and an output holding torque can be generated. However, the comb jamming mechanism uses the bending of thin plates and contact of the plates according to the gain of normal load. Although the load towards layer is small, a holding torque is not generated; given that the friction is excessively low for the joint to be fixed.

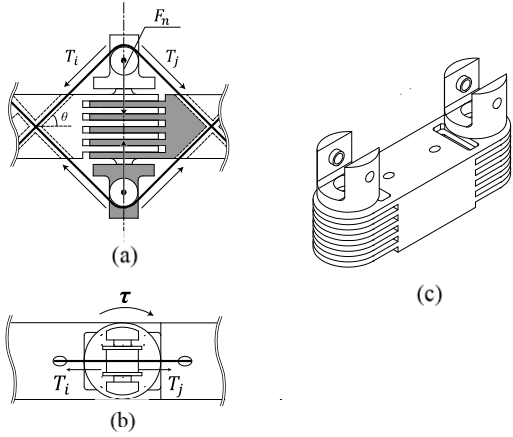

Fig. 5. Comb jamming: (a) Principle of side-view and (b) principle of top-view, and (c) perspective view

### C. Radial Layer Jamming

Thin-plate laminates were processed using the radial-layer jamming mechanism along the r-axis, as shown in Fig. 2. The joints were placed within thin cylinders in the shapes of concentric circles with a specified clearance distance, and the wire was positioned at the center of the part. The string configuration was composed by combining the thin cylinders (Fig. 3). Numerous researches were conducted on multi cylindrical devices [15]. However, they are based on Magneto Rheological Fluid (MRF), which is nonlinear. Moreover, a tendon drive was employed, and a string configuration was composed using wire. Furthermore, as mentioned previously, this is a novel layer jamming mechanism.

The wire runs through the center of the parts, which is the same for the bead jamming mechanism. Therefore, the string configuration can maintain its shape by the application of a tensile load, given that the path length of the wire remains almost change. By the pulling wire, the thin cylinders are subjected to compressive loads and bending. Hence, they come into contact, and friction is generated. When an attempt is made to rotate the joint, a reaction moment is generated due to friction. In this paper, it was defined as the holding torque.

It was hypothesized that the thin cylinder does not bend; thus, the theoretical value of holding torque was used. With focus on several thin cylinders, the contact condition was found to be the same as that in bead jamming. Hence, the holding torque should be represented $\tau_i = \mu R_i T$, where the length from the joint center to the contact area is $R_i$. This is in accordance with the following definition: the total holding torque is the addition of the torques of each cylindrical radius. Moreover, it can be expressed as (6):

$$\tau_{\text{radial}} = \sum_i \mu R_i T \qquad (6)$$

The string configuration developed with this mechanism can bend significantly by shortening the wheel-base of a part, as shown in Fig. 6.

The radial-layer jamming mechanism is superior to the comb jamming mechanism in that the holding torque does not decrease toward the tip of the string configuration. This is because there is no rolling friction, due to its composition. With respect to the 3D bending of the string configuration, a part with a yaw-axis and pitch-axis as joints was composed, in accordance with the method. Comb jamming mechanism needs to train wire cross each other in such a way and it makes difficult to bending 3D due to interference, and the range of motion of its joints are limited. This does not occur in the radial-layer jamming mechanism, which is an advantage when compared with the comb jamming mechanism.

A part with a shortened wheel base was developed, which had a male and female structure. The string configuration was composed using these structures, and an experiment was conducted.

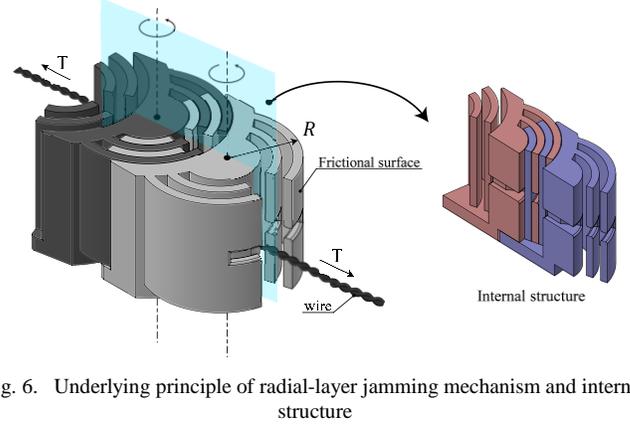

Fig. 6. Underlying principle of radial-layer jamming mechanism and internal structure

### III. DESIGN CONCEPT

Prototypes were developed using the bead, comb and radial-layer jamming mechanisms, as discussed in Section II (Fig. 1 corresponds to the bead jamming mechanism; and Figs. 7(a) and (b) correspond to the comb and radial-layer jamming mechanisms, respectively). Moreover, the prototype specifications are listed in TABLE I.

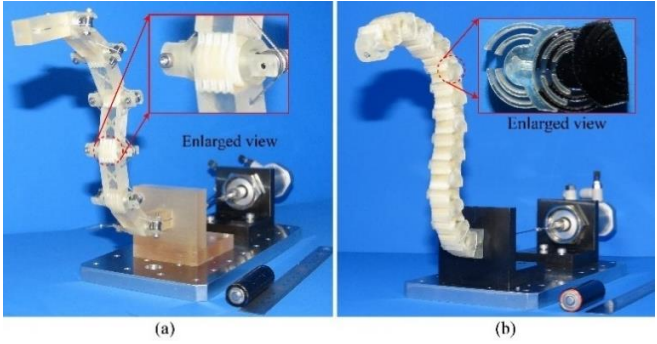

Fig. 7. Prototypes developed using the (a) comb and (b) radial-layer jamming mechanisms

TABLE I. SPECIFICATIONS OF A PART

|  | Beads | Comb | Radial |
|---|---|---|---|
| **Width (mm)** | 12.0 | 12.0 | 15.0 |
| **Height (mm)** | 11.0 | 33.0 | 13.5 |
| **Inter-axial distance (mm)** | 6 | 30.0 | 9.1 |
| **Weight (g)** | 0.74 | 7.27 | 1.40 |
| **Number of plates** |  | 5 | 3 |
| **Thickness of plate (mm)** |  | 1.10 | 0.90 |
| **Clearance distance (mm)** |  | 1.12 | 1.05 |

### A. Design Considerations

In this study, three prototypes were developed using three types of mechanisms. The diameters of the joints were approximately 10–15 mm, given that the experiment was conducted at the same scale.

The bead jamming mechanism can only realize two-dimensional (2D) bending and motion, as shown in Fig. 4. However, 3D motion can be realized by making the spherical surface the contact area. In this study, the cone angle was set as 70° for the wire route. The curves inside the beads were smoothed to decrease the friction between the wire and inner-wall. Moreover, wire with a breaking load of 19.9 kgf was used.

In the comb jamming mechanism, the bearing is used as the pulley for the addition of a normal load to the layer structure. The wheel-base of the radial layer jamming mechanism is dependent on the angle $\theta$ of the angular training wire; thus, $\theta = 45°$ was set as a representative figure. A through-hole was added to the base material in proportion to $\theta$, and five plates were used. The thin-plate beam was perforated at its root, to allow for easy bending. To prevent the deviation of the plate position, a thin-walled pipe was placed inserted into the through-hole made in the layers. Moreover, the pulley is also stabilized at the thin-walled pipe, and it can transmit loads in the direction of the layer.

Two strategies were employed for the prototype developed using the radial-layer jamming mechanism. First, the layers contained slits for the transmission of compressive loads to the center of the layer structure. Moreover, there were slits along the z-axis, and the compressive load could be transmitted through the wire. Second, for the easy reception of compressive loads, the thin-walled pipes were made to overhang in the r-axis. This is because compressive loads are applied along r-axis. These realize that thin cylinder quickly contact with each other and holding torque occurs with low tensile load.

### B. Units

- To make the same width of three types 15 mm, comb mechanism's layer 5 and radial layer mechanism 's one is 3.
- The manner of bending is different between the simple thin plate and thin cylinder. So, their plate's thickness and clearance between them are also different. In the comb jamming mechanism, the thickness of the plate was 1.10 mm, and the clearance was 1.12 mm. In the radial-layer mechanism, the thickness of the plate was 0.90 mm, and the clearance was 1.05 mm. Moreover, these values were empirically determined.

The parts were made from acrylic resin (AR-M2), and the parts were developed using stereolithography (Keyence, AGILISTA-3100; Osaka, Japan).

## IV. EXPERIMENTS

Two experiments were conducted, and the holding torques for the three types of mechanisms were measured using a tension tester (INSTRON 3343, load capacity; 1 kN).

(i) The relationship between the number joints and the average holding torque.

### A. Measuring Method

The experimental conditions are shown in Fig. 8, and the experimental procedure was carried out as follows:.

*1)* Air is transferred to the pressurizing cylinder, and the wire connected to it is then pulled; which initiates the jamming phenomenon

*2)* The end of the string configuration is pushed until the maximum value is measured.

*3)* The maximum value as a reaction force measured by pushing is considered as the holding force. The value multiplied by the length from the center of the joints to the end of string configurations is then considered as the maximum holding torque.

*4)* By repeating the abovementioned steps for each joint, the maximum holding torque was measured, and the average values were considered.

The conditions of the experiment were as follows:

- The length from the center of a target joint to the end of the string configuration was 0.122 m, and it was pushed by a total distance of 15 mm.

- The velocity of pushing was 10 mm/min.
- Mounts were developed for the stabilizing of the target joint to be measured.
- An accurate regulator was used for the control of the tensile load, and its value was 50 N.

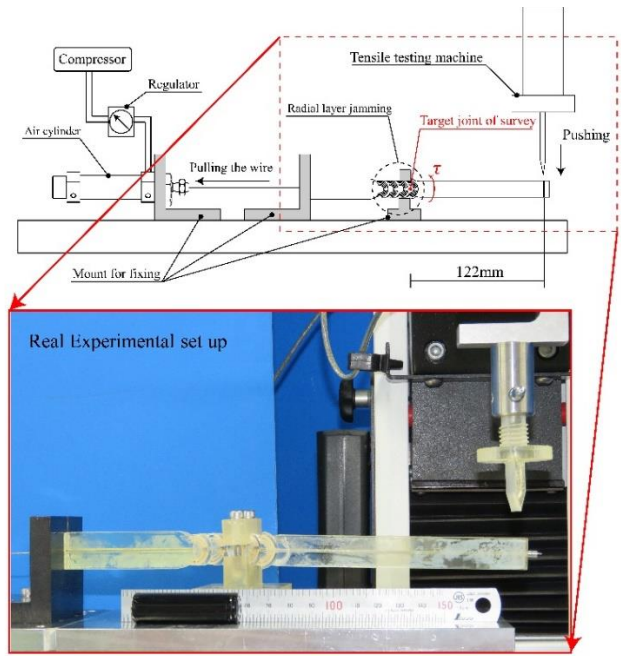

Fig. 8. Experimental setup for evaluation of average holding torque

### B. Evaluation Method and Considerations

The experimental results are shown in Fig. 9. The average holding torque of each joint was measured, and then compared.

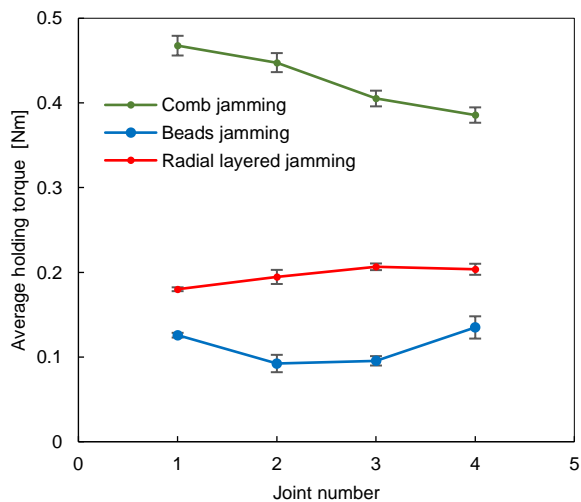

Fig. 9. (i) Experimental results

The holding torques of the bead and radial-layer jamming mechanisms were found to be equal in all joints; however, these differed from that of comb jamming mechanism. Given that its holding torque decreased towards the tip of string configurations, in accordance with its underlying principle, this experiment was conducted. In particular, the holding torque decreased as shown in Fig. 9. The comb jamming mechanism uses the pulley for the transmission of the tensile load and the determination of its direction. Moreover, rolling friction is generated. The holding torque was found to decrease by 6% per joint; which is greater than the decrease rates of the bead jamming and radial-layer mechanism by factors of 4 and 2, respectively. Compared with the comb jamming mechanism, the torques of the bead jamming and radial-layer jamming mechanisms did not decrease gradually, although the values have some extent dispersions. The fourth holding torques of the bead jamming, comb jamming, and radial-layer mechanisms were 0.13 Nm, 0.20 Nm, and 0.39 Nm, respectively. If the string configuration has 10 joints, the torque of its tip is halved; thus, the torque of the radial-layer jamming mechanism exceeded that of the comb jamming mechanism. The radial-layer jamming mechanism can realize long string configurations, and its holding torques are almost equal at all joints. Moreover, the radial-layer mechanism was found to be superior to the bead jamming and comb jamming mechanisms in these respects.

The material characteristics are related to the fact that the first and fourth torques of the bead jamming mechanism are higher than the others, and that of the radial-layer jamming mechanism gradually increases. The surface of its material was excessively adhesive for the 3D printer characteristics. Given that the surfaces easily adhered to each other, slight derricking was observed, and the surfaces were combined. Therefore, the clearance was in a vacuum-state such as glass, and a resistance force different from friction was generated. Moreover, it was difficult to measure its frictional coefficient, as the occurrence of the phenomenon is complicated [16] [17]. considerate was concluded that the differences between the holding torques were due to the indeterminate surface properties of the joints.

(ii) The relationship between tensile load and average holding torque

### A. Measuring Method

Steps 1)–3) were the same as that in Experiment (i). In this experiment, only the first holding torques of the joints were measured.

4) The holding torque of each joints was measured by repeating Steps 1)–3).

### B. Evaluation Method and Considerations

The experimental results are shown in Fig. 10.

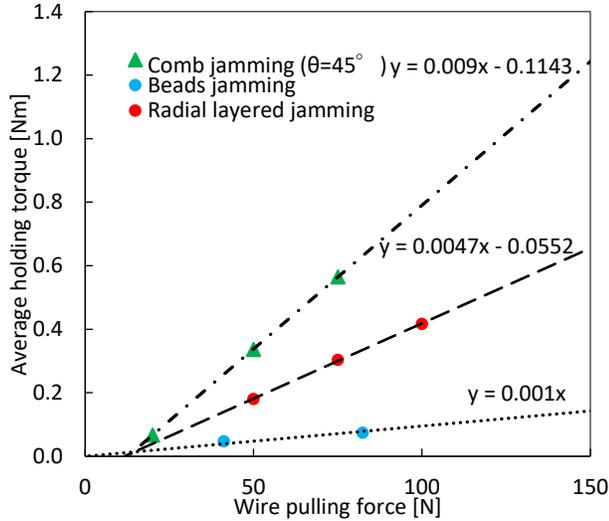

Fig. 10. Results of Experiment (ii)

From the experimental results, the holding torque was found to increase in accordance with an increase in the tensile load. Moreover, the training angle was set in the range of $0° \leq \theta \leq 90°$; thus, $\theta = 45°$ was selected as the representative value. In this case, the contact angle between the wire and pulley was 90°.

The comb jamming mechanism exhibited the most significant increase in holding torques. The bead jamming mechanism does not involve layers, and the holding torque was found to increase in accordance with an increase in the frictional coefficient. The radial-layer mechanism is in-between these two types; as shown in Fig. 11. Moreover, the focus was on the radius of the layer.

The maximum holding torques of the comb jamming mechanism are experienced by each plate, and the total holding torque is dependent on the number of contact areas. The number of contact areas can be expressed as $2a - 1$, where $a$ is the number of contact surfaces (including innermost wall). Therefore, the comb jamming mechanism achieves maximum torque in each plate; whereas, the radial-layer mechanism exhibits a torque in the outermost cylindrical surface, as shown in Fig. 11. In summary, their abundance pattern is different, given that the average radius of layer is different when comparing the comb and radial-layer mechanisms.

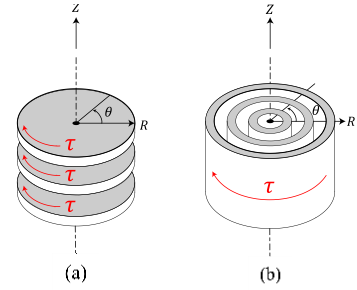

Fig. 11. The difference between the average radii of comb jamming and radial-layer mechanisms

The intercept of the radial-layer jamming mechanism was found to be higher than that of the comb jamming mechanism; which indicates that the radial-layer jamming mechanism requires a lower tensile load than the comb jamming mechanism, although the thin cylinder of the radial layer is difficult to bend when compared with the thin plate of the comb jamming mechanism. This is because there were slits in the thin cylinder along z-axis, thus rendering it easy to bend. Without the slits, the minimum tensile load that generates the torque would be excessively large, given that the thin cylinder requires a large stress for bending to be realized, when compared with the bending of the thin-plate beam. Furthermore, the slit-feature has an influence on the responsiveness of the mechanism with respect to tensile loads.

## V. CONCLUSIONS

In this paper, the following three types of jamming mechanisms with tunable stiffnesses are proposed: the bead jamming, comb jamming, and radial-layer jamming mechanisms. The mechanisms and designs were developed. Moreover, two experiments were conducted on the developed prototypes, and the experimental results were obtained. The radial-layer jamming mechanism is in-between the bead jamming and comb jamming mechanisms with respect to the holding torque. The holding torque maintains its value if the string configuration comes is excessively long, as confirmed by Experiment (i); thus demonstrating the relationship between the number of joints and the average holding torque. The holding torque of the radial-layer mechanism was found to be greater than that of the bead jamming mechanism considering the intake of the layer structure between the tensile load and the average holding torque, as confirmed by Experiment (ii).

The string configuration of the radial-layer mechanism was found to be flexible, and it exhibited a constant holding torque, which is similar to the bead jamming mechanism. Moreover, the holding torque was improved by employing the layer structure in its joints. Moreover, a novel layer jamming mechanism can be developed using wire that combines the advantages of the bead jamming and comb jamming mechanisms. In future work, accurate theoretical values will be obtained for the three types of mechanisms by conducting a finite element analysis under the consideration of the bending of the material, due to the complex shapes resulting from the mechanisms. Moreover, the most suitable design method that corresponds to each string configuration length from the wheel-base of a part and movable range of a joint will be implemented. In addition, significant research attention will be directed toward the development of new applications, grippers, and lock mechanisms, among other schemes.

ACKNOWLEDGMENT